\documentclass[11pt,a4paper]{article}
\usepackage[hyperref]{emnlp2020}
\usepackage{times}
\usepackage{latexsym}

\usepackage{microtype}
\usepackage{graphicx} % for inserting the image

\aclfinalcopy % Uncomment this line for the final submission
\usepackage{array}
\newcolumntype{M}[1]{>{\centering\let\newline\\\arraybackslash\hspace{0pt}}m{#1}}
\newcolumntype{L}[1]{>{\centering\let\newline\\\arraybackslash\hspace{0pt}}m{#1}}
\newcolumntype{C}[1]{>{\centering\let\newline\\\arraybackslash\hspace{0pt}}m{#1}}
\newcolumntype{R}[1]{>{\centering\let\newline\\\arraybackslash\hspace{0pt}}m{#1}}
\newcolumntype{P}[1]{>{\centering\let\newline\\\arraybackslash\hspace{0pt}}m{#1}}

\definecolor{KDpurple}{rgb}{0.6,0.18,0.64}

\title{Benchmarking BioRelEx for Entity Tagging and Relation Extraction}

\author{Abhinav Bhatt \\
    Amelia Science \\
    RnD, IPsoft\\
    Bengaluru, KA 560066\\
  \texttt{abhatt@ipsoft.com} \\\And
  Kaustubh D. Dhole \\
   Amelia Science \\
    RnD, IPsoft\\
    New York, NY 10004\\
  \texttt{kdhole@ipsoft.com} \\}

\date{}

\begin{document}
\maketitle
\begin{abstract}
Extracting relationships and interactions between different biological entities is still an extremely challenging problem but has not received much attention as much as extraction in other generic domains. In addition to the lack of annotated data, low benchmarking is still a major reason for slow progress. In order to fill this gap, we compare multiple existing entity and relation extraction models over a recently introduced public dataset, BioRelEx of sentences annotated with biological entities and relations. Our straightforward benchmarking shows that span-based multi-task architectures like DYGIE show 4.9\% and 6\% absolute improvements in entity tagging and relation extraction respectively over the previous state-of-art and that incorporating domain-specific information like embeddings pre-trained over related domains boosts performance.
\end{abstract}

\section{Introduction}
Extracting biological entities and determining relationships between them is particularly challenging due to the heterogeneous nomenclature of such entities. Besides, alongwith the rapid changes in molecular biology, there is a lack of consensus over a standardized dataset.
While~\citet{khachatrian-etal-2019-biorelex} attempt to address the issue of standardization by introducing BioRelEx, there is still plenty of room for improving the tagging and relation extraction of these entities. 

Especially when there has been tremendous progress achieved by multi-task span-based architectures to extract entities and relations in various datasets within domains like news and scientific articles~\cite{Luan_2019, wadden-etal-2019-entity}, we attempt to seek if this success can be emulated to recent standardized biological literature, like BioRelEx. 

We compare a range of classifiers; previously common generic ones like CRF and BiLSTM as well as recent state-of-the-art (SOA) multi-task span based architectures like Dynamic Graph IE (DYGIE)~\cite{Luan_2018, Luan_2019} in order to provide a stronger benchmark over BioRelEx~\cite{khachatrian-etal-2019-biorelex}, a recently published dataset of 2010 sentences consisting annotations of biological entities and binding interactions between those entities. We also check the impacts of using DYGIE's updated span representations obtained from relation and coreference graph propagation. Given the domain specificity of BioRelEx, we assess the impact of utilizing embeddings pre-trained on related domains.

\section{Related Work}
\begin{figure}
    \centering
    \includegraphics[scale=0.2]{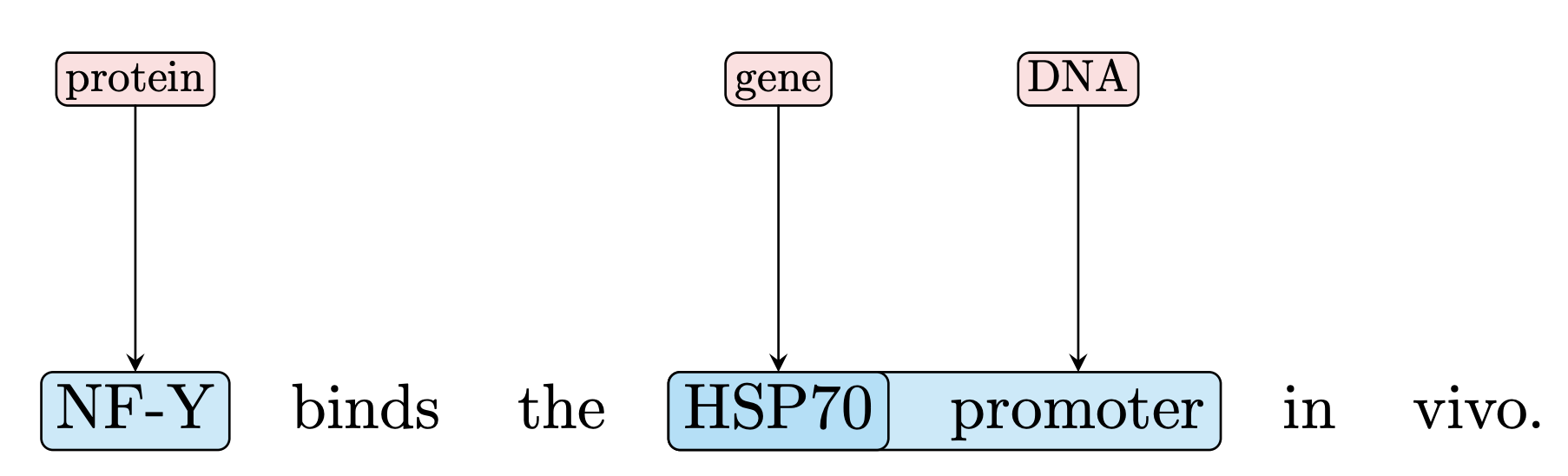}
    \caption{BioRelEx has 33 types of entities including annotations of multiple overlapping entities. (Entity Tagging)}
    \label{fig:fig_et}
\end{figure}
\begin{figure}
    \centering
    \includegraphics[scale=0.2]{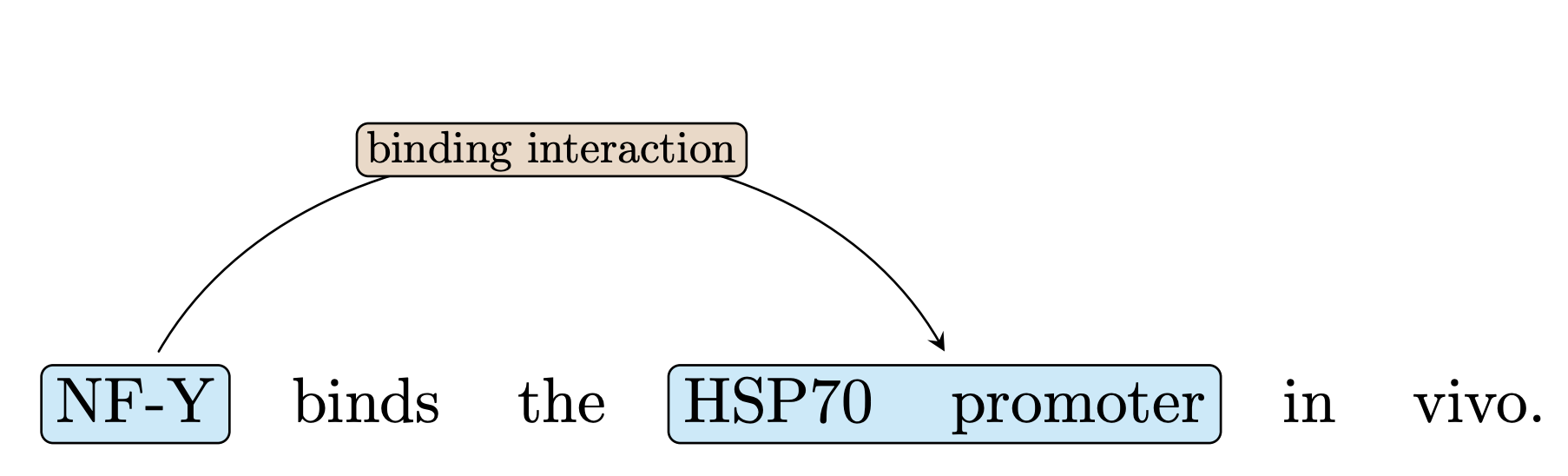}
    \caption{A sample Binding Interaction in BioRelEx between two BioRelEx entities (Relation Extraction)}
    \label{fig:fig_re}
\end{figure}
Most information extraction systems have looked at entity tagging as a flat sequence labeling problem and have employed feature-based CRFs and later neural architectures~\cite{ma-hovy-2016-end, peters2018deep}.~\citet{lample-etal-2016-neural} take inspiration from shift-reduce parsers to employ a transition based approach.
~\citet{yu2010jointly, miwa2014modeling, li-ji-2014-incremental, Zheng_2017} have jointly modelled entity tagging and relation extraction.~\citet{fu2019graphrel} use relation-weighted graph convolution networks for considering the interaction between named entities and relations.

To incorporate the ability to tag overlapping spans,~\citet{katiyar2018nested, wang2018neural, Wang_2019} devised neural methods applying hyper graph based representations on top of sequence labelling.

~\citet{8982966, dai2020effective} tag nested and discontinuous mentions in biomedical text.~\citet{Lee_2017, lee-etal-2018-higher} enumerate over all possible spans and apply bean pruning to reduce the number of candidate spans used for coreference resolution.~\citet{Luan_2018, Luan_2019} further apply a multi-task graph based architecture incorporating relation, coreference and event based propagation to enhance span representations. 

~\citet{lin2020rigourous} test how well do current NER models generalize in the absence of strong name regularity, high mention coverage and sufficient context diversity.

~\citet{khachatrian-etal-2019-biorelex} introduce BioRelEx by annotating bio-medical texts with entities grounded with other databases, focusing on delineating biological and experimental entities and distinguishing between direct regulatory and indirect physical interactions.
% 1. Entity Extraction/Relation Extraction Review
% 2. BioRelex review

\section{Dataset \& Evaluation}
~\citet{khachatrian-etal-2019-biorelex} have made publicly available 2010 annotated sentences (1405 for training and 201 for validation) consisting of annotations for 33 types of entities  and relation annotations for binding interactions. We gauge the performance of all architectures over the validation dataset using the evaluation script of ~\citet{khachatrian-etal-2019-biorelex} namely measuring the micro-averaged precision, recall and F-score.

\section{Entity Tagging}
Entity Tagging (Figure~\ref{fig:fig_et}) is the task of predicting the best entity type label for every span of words in a given piece of text. And Relation Extraction (Figure~\ref{fig:fig_re}) involves predicting the best relation type for such entity spans.
\subsection{Traditional Architectures}
% Huang https://arxiv.org/abs/1508.01991

We evaluate various architectures traditionally common for the task of entity tagging which have displayed strong performances on other entity tagging datasets. As a baseline, we train a BiLSTM-CRF~\cite{huang2015bidirectional}, a BiGRU-CRF and a CNN-CRF~\cite{10.5555/1953048.2078186}
with 300-d Glove embeddings~\cite{pennington-etal-2014-glove} as input. 

BioRelEx, like other bio-medical texts, belong to a knowledge intensive domain and hence we expect a considerable amount of domain-specific terms to appear. In accordance with that, we incorporate sub-word embeddings by taking advantage of
\begin{itemize}
\itemsep0em 
\item Flair's~\cite{akbik2019flair} implementation of Contextual String Embeddings~\cite{akbik2018coling} which use a character-language model
\item Byte Pair Embeddings~\cite{heinzerling-strube-2018-bpemb}
\item FastText Embeddings~\cite{bojanowski-etal-2017-enriching}
\end{itemize}
as inputs to a BiLSTM. Table~\ref{flat-arch-table} compares the performance of these traditional architectures alongwith the various embeddings on BioRelEx's validation dataset. Our experiments show that Flair embeddings provide a clear gain of 14\% over other embeddings. 
However, flat architectures like BiLSTM unarguably lack the ability to model overlapping spans and hence in the next section, we attempt to use recently successful architectures like DYGIE to check for performance gains.

\begin{table*}[!hbpt]
\centering
\small
\begin{tabular}
    {| >{\raggedright}m{6.5cm}
   | >{\raggedright}m{2cm}
   | >{\raggedright}m{2cm}
   | >{\raggedright\arraybackslash}m{2cm}|}
   \hline \textbf{System} & \textbf{Precision} & \textbf{Recall} & \textbf{F-score} \\ \hline
BiLSTM-CRF (Glove) & 0.516 & 0.525 & 0.520 \\
BiGRU-CRF (Glove) & 0.508 & 0.483 & 0.495 \\
CNN-CRF (Glove) & 0.541 & 0.500 & 0.520 \\
% _{news-forward + news-backward + glove}
\hline
News-forward + News-backward + Glove & 0.723 & 0.604 & 0.658 \\
% _{Pubmed-forward + Pubmed-backward + BytePairEmbeddings}
Pubmed-forward + Pubmed-backward + BytePairEmbeddings & 0.743 & 0.596 & 0.662 \\
% _{Pubmed-forward + Pubmed-backward + BytePairEmbeddings + FastText}
Pubmed-forward + Pubmed-backward + BytePairEmbeddings + FastText & 0.748 & 0.591 & \textbf{0.660} \\
\hline
\end{tabular}
\caption{Performance of Flat Architectures on Entity Tagging. News-* and Pubmed-* are character level embeddings~\cite{akbik2018coling, akbik2019flair} used as inputs to a BiLSTM-CRF.}
\label{flat-arch-table}
\end{table*}

\begin{table*}[!hbpt]
\centering
\small
\begin{tabular}
    {| >{\raggedright}m{6.5cm}
   | >{\raggedright}m{2cm}
   | >{\raggedright}m{2cm}
   | >{\raggedright\arraybackslash}m{2cm}|}
   \hline \textbf{System} & \textbf{Precision} & \textbf{Recall} & \textbf{F-score} \\ \hline
DYGIEPP (SciBERT) & 0.587 & 0.630 & 0.600 (-0.208)\\ %{SciBert, no coref propagation, no relation_propagation, 400 units, 3 layers}
DYGIEPP (BERT-base) & 0.567 & 0.660 & 0.610 (-0.198)\\ 
%{Bert Base, no coref propagation, no relation_propagation, 400 units, 3 layers}
% DYGIEPP def elmo, no coref, no rel, 400 units, 3 layers
\hline
DYGIEPP & 0.832 & 0.786 & 0.808 \\ %{no coref propagation, no relation_propagation, 400 units, 3 layers}
+Relation Propagation (RP) & 0.845 & 0.766 & 0.803 (-0.005) \\ %{no coref propagation, relation_propagation, 400 units, 3 layers}
+Coreference Propagation (CP) & 0.837 & 0.787 & \textbf{0.811} (+0.003) \\
+RP, CP & 0.832 & 0.767 & 0.798 (-0.010) \\
DYGIEPP (200 BiLSTM units 1 layer) & 0.836 & 0.769 & 0.801 (-0.007) \\ %{no coref propagation, no relation_propagation, 200 units, 1 layer}
DYGIEPP (400 BiLSTM units 2 layers) & 0.826 & 0.773 & 0.800 (-0.008) \\ %{no coref propagation, no relation_propagation, 200 units, 1 layer}
DYGIEPP (512 BiLSTM units 3 layers) & 0.852 & 0.771 & 0.810 (+0.002) \\ %{no coref propagation, no relation_propagation, 512 units, 3 layers}
\hline
SciERC & 0.864 & 0.710 & 0.779 \\
\hline
\end{tabular}
\caption{Performance of Multi-Task Architectures on Entity Tagging. DYGIEPP mentioned in the third row uses ELMo embeddings but without graph propagation. Also it has 400 BiLSTM units and 3 BiLSTM layers.}
\label{multitask-arch-table}
\end{table*}

\begin{table*}
\centering
\small
\begin{tabular}
    {| >{\raggedright}m{6.5cm}
   | >{\raggedright}m{2cm}
   | >{\raggedright}m{2cm}
   | >{\raggedright\arraybackslash}m{2cm}|}
   \hline \textbf{System} & \textbf{Precision} & \textbf{Recall} & \textbf{F-score} \\ \hline
% default %{Elmo, Glove, no coref propagation, no relation propagation, 400 units, 3 layers}
+BioELMo & 0.858 & 0.790 & 0.823 (+0.015) \\ %{BioElmo, no coref propagation, no relation_propagation, 400 units, 3 layers}
+BioELMo, Relation Propagation (RP) & 0.862 & 0.794 & 0.827 (+0.019) \\ %{BioElmo, no coref propagation, relation_propagation, 400 units, 3 layers}
+BioELMo, Coreference Propagation (CP) & 0.858 & 0.792 & 0.824 (+0.016)\\ %{BioElmo, coref propagation, no relation propagation, 400 units, 3 layers}
+BioELMo, RP, CP & 0.851 & 0.805 & \textbf{0.828} (+0.020)\\ %{BioElmo, coref propagation, relation_propagation, 400 units, 3 layers}
+BioELMo, 200 BiLSTM units 1 layer & 0.836 & 0.807 & 0.821 (+0.013)\\ %{BioElmo, no coref propagation, no relation propagation, 200 units, 1 layer}
+BioELMo, 400 BiLSTM units 2 layers & 0.859 & 0.790 & 0.822 (+0.014)\\ %{BioElmo, no coref propagation, no relation propagation, 400 units, 2 layers}
+BioELMo, BioWord2Vec & 0.844 & 0.793 & 0.818 (+0.010)\\ %{BioElmo, BioWord2Vec, no coref propagation, no relation propagation, 400 units, 3 layers}
\hline
\end{tabular}
\caption{Impact of using Domain Specific Embeddings on Entity Tagging. We perform the experiments on DYGIEPP with the same base configuration as that of the Section 4.2 (Table~\ref{multitask-arch-table})}
\label{domain-specific-embeddings-table}
\end{table*}
\begin{table*}
\centering
\small
\begin{tabular}
    {| >{\raggedright}m{6.5cm}
   | >{\raggedright}m{2cm}
   | >{\raggedright}m{2cm}
   | >{\raggedright\arraybackslash}m{2cm}|}
\hline \textbf{System} & \textbf{Precision} & \textbf{Recall} & \textbf{F-score} \\ \hline
DYGIEPP (SciBERT) & 0.241 & 0.139 & 0.176 (-0.379)\\ %{SciBert, no coref propagation, no relation_propagation, 400 units, 3 layers}
DYGIEPP (BERT-base) & 0.239 & 0.109 & 0.150 (-0.405)\\ 
%{Bert Base, no coref propagation, no relation_propagation, 400 units, 3 layers}
% Base same as multitask for entity tagging
\hline
DYGIEPP & 0.592 & 0.523 & 0.555  \\ %{no coref propagation, no relation_propagation, 400 units, 3 layers}
+Relation Propagation (RP) & 0.628 & 0.500 & \textbf{0.556} (+0.001) \\ %{no coref propagation, relation_propagation, 400 units, 3 layers}
+Coreference Propagation (CP) & 0.450 & 0.520 & 0.482 (-0.073) \\
+RP, CP & 0.479 & 0.527 & 0.502 (-0.053) \\
DYGIEPP (200 BiLSTM units 1 layer) & 0.569 & 0.459 & 0.508 (-0.047)\\ %{no coref propagation, no relation_propagation, 200 units, 1 layer}
DYGIEPP (400 BiLSTM units 2 layers) & 0.570 & 0.452 & 0.504 (-0.051) \\ %{no coref propagation, no relation_propagation, 200 units, 1 layer}
DYGIEPP (512 BiLSTM units 3 layers) & 0.594 & 0.503 & 0.545 (-0.010) \\ %{no coref propagation, no relation_propagation, 512 units, 3 layers}
\hline
SciERC & 0.490 & 0.503 & 0.496 \\
\hline
\end{tabular}
\caption{Relation Extraction Performance of Multi-Task Architectures. We use the same base configuration of DYGIEPP as the one in Section 4.2 (Table~\ref{multitask-arch-table})}
\label{multitask-arch-table-relations} 
\end{table*}

\begin{table*}
\centering
\small
\begin{tabular}
    {| >{\raggedright}m{6.5cm}
   | >{\raggedright}m{2cm}
   | >{\raggedright}m{2cm}
   | >{\raggedright\arraybackslash}m{2cm}|}
\hline \textbf{System} & \textbf{Precision} & \textbf{Recall} & \textbf{F-score} \\ \hline
+BioELMo & 0.584 & 0.517 & 0.548 (-0.007)\\ %{BioElmo, no coref propagation, no relation_propagation, 400 units, 3 layers}
+BioELMo, Relation Propagation (RP) & 0.571 & 0.520 & 0.544 (-0.011) \\ %{BioElmo, no coref propagation, relation_propagation, 400 units, 3 layers}
+BioELMo, Coreference Propagation (CP) & 0.498 & 0.568 & 0.531 (-0.024) \\ %{BioElmo, coref propagation, no relation propagation, 400 units, 3 layers}
+BioELMo, RP, CP & 0.424 & 0.557 & 0.482 (-0.073)\\ %{BioElmo, coref propagation, relation_propagation, 400 units, 3 layers}
+BioELMo, 200 BiLSTM units 1 layer & 0.469 & 0.557 & 0.510 (-0.045)\\ %{BioElmo, no coref propagation, no relation propagation, 200 units, 1 layer}
+BioELMo, 400 BiLSTM units 2 layers & 0.582 & 0.503 & 0.540 (-0.015)\\ %{BioElmo, no coref propagation, no relation propagation, 200 units, 1 layer}
+BioELMo, BioWord2Vec & 0.521 & 0.493 & 0.506 (-0.049) \\ %{BioElmo, BioWord2Vec, no coref propagation, no relation propagation, 400 units, 3 layers}
\hline
\end{tabular}
\caption{Relation Extraction Performance of Domain Specific Embeddings over DYGIEPP }
\label{domain-specific-embeddings-table-relations}
\end{table*}

\subsection{Multi-Task Architectures}
We choose to evaluate DYGIEPP~\cite{wadden-etal-2019-entity} which has shown strong results on many entity tagging datasets and emphasized the importance of pre-training with text from similar domains like SciBERT~\cite{Beltagy2019SciBERT}.

In Table~\ref{multitask-arch-table}, we present different configurations of DYGIEPP alongwith their results. Surprisingly, concatenation of ELMo Embeddings~\cite{peters2018deep}, Glove and Character Embeddings performs better than SciBERT for the BioRelEx dataset. Although, relation propagation do not seem to help much, coreference propagation shows an increase in the scores. Moreover, we observe that increasing the number of hidden units and layers boosts the scores.~\footnote{We find that keeping 400 hidden units and 2 layers in the BiLSTM works best for our case. So, we use this configuration in all our future experiments.} 

DYGIEPP's best configuration performs 3.2\% better than the current SOA, SciERC~\cite{Beltagy2019SciBERT, khachatrian-etal-2019-biorelex}.

\subsection{Domain-Specific Embeddings}
Domain-specific embeddings have been essential to achieve high performances for entity tagging tasks especially in knowledge intensive domains.~\cite{romanov2018lessons, zhu2018clinical}.
Since BioRelEx is a domain specific dataset, we experimented with embeddings trained on related domains as well as with embeddings pre-trained on BioRelEx. 

We replaced the original ELMo embeddings in~\citet{wadden-etal-2019-entity} with BioELMo~\cite{jin2019probing} which have been pre-trained on 10M PubMed abstracts. We use Word2Vec embeddings trained on PubMed data~\cite{Pyysalo2013DistributionalSR} too, which we will refer to as BioWord2Vec and also train Glove embeddings on the BioRelEx dataset. We present the results in Table~\ref{domain-specific-embeddings-table}. We observe that incorporating BioELMo boosts performance of DYGIEPP by an absolute 1.7\% over ELMo embeddings. Using coreference propagation and relation propagation reveal slight improvement as well. 

\section{Relation Extraction}

\subsection{Multi-Task Architectures}
Since the multi-task architecture proposed by~\citet{Luan_2018,wadden-etal-2019-entity} have attempted to leverage context across multiple sentences as well as incorporate coreference, relation and event information to enhance span representations, we verify if this success would easily apply to BioRelEx since this dataset contains a notable amount of coreferences and biologically nested entities like shown in Figure~\ref{fig:fig_et}. 

We summarise the results for relation extraction in Table~\ref{multitask-arch-table-relations}. Here, the combination of ELMo, Glove and Character Embeddings performs remarkably better than SciBERT. Contrary to entity tagging, for relation extraction, enabling relation propagation gives us the best scores. Further increasing the layers and the number of units in the BiLSTM also helps a lot in this case.

\subsection{Domain-Specific Embeddings}

We follow the same steps as followed for entity tagging. We present the results for relation extraction in Table~\ref{domain-specific-embeddings-table-relations}. We observe that whereas in case of entity tagging, using domain specific embeddings, specifically BioELMo shows a good improvement, for relation extraction, they show a slight decrease in the F-score. Comparing Table~\ref{domain-specific-embeddings-table-relations} to Table~\ref{multitask-arch-table-relations}, we see a small reduction when using domain specific embeddings. Similar to entity tagging, here too, BioWord2Vec reduces the F-score.

\section{Conclusion}
We evaluate a number of architectures for entity tagging and relation extraction and perform extensive hyperparameter tuning over the BioRelEx dataset. Our experiments improve BioRelEx's SOA for entity tagging and relation extraction by 4.9\% and 6\% F-score respectively. 
Our evaluations show that using ELMo, Glove and character embeddings improves the performance compared to BERT and SciBERT on multi-task architectures. Likewise, utilizing domain specific information in the form of pre-trained embeddings also leads to an improvement in performance for entity tagging.

\section{Discussion}
Multi-task architectures like DYGIE, by and large with domain-specific embeddings improve the predictions of BioRelEx. Interestingly, we find that contextual embeddings like ELMo and BioELMo exceed the performance of BERT~\cite{devlin2018bert} and SciBERT by a large margin which is tangential to findings on other benchmarks like GLUE~\cite{wang-etal-2018-glue}. 

%One thing to note is that while training DYGIEPP or SciERC models, the data needs to be converted in their format from the format in which \citet{khachatrian-etal-2019-biorelex} have given us. And since the scores which we have presented are from the evaluation script given by \citet{khachatrian-etal-2019-biorelex}, we have to convert the data from the predicted format to the BioRelEx format again. 
 Besides, we observe that in a few cases, certain span annotations of entities are not being taken into account while converting to SciERC's input format by the provided conversion script, but were being expected to be predicted during the evaluation process making it impossible for any model to go beyond a training set F-score of 91.2\%.\footnote{When we convert the dev set using the provided conversion script, the F-score from the evaluation script gives us 0.915 F-score for entity tagging and 0.834 for relation extraction, while on the training set it goes to 0.912 for entity tagging and 0.850 for relation extraction.} While this minor incorporation in BioRelEx might drastically improve F-scores for all of the architectures, relative differences of performances between the architectures might hardly change. We would still expect multi-task architectures to predict BioRelEx entities and relations better vis-à-vis traditional architectures as the F-scores display an appreciable leap of 15\% to 30\% already (Table~\ref{flat-arch-table} vs Table~\ref{multitask-arch-table}).
% Dev
% Entities - 0.915 
% Rel_ext - 0.834

% Train
% Etities - 0.912
% Rel_ext - 0.850

\bibliography{anthology,emnlp2020}
\bibliographystyle{acl_natbib}

\end{document}